\documentclass[10pt,twocolumn,letterpaper]{article}

\usepackage{color}
\usepackage{xcolor}
\usepackage{graphicx}
\usepackage{float}
\usepackage{xspace}
\usepackage{lipsum}
\usepackage[listofformat=parens, justification=centering, subrefformat=subparens]{subfig}
\usepackage{amsmath,amssymb}
\usepackage{titlesec}
\usepackage{tabularx}

\usepackage{framed,multirow,tabularx}
\usepackage[point]{fltpoint}
\usepackage{todonotes}
\usepackage{textcomp}
\usepackage{ifthen}
\usepackage{latexsym}
\usepackage{fontawesome5} 
\usepackage{titling}

\usepackage{tikz}
\usetikzlibrary{positioning, shapes.geometric, arrows.meta, calc}

\usepackage{pifont}
\DeclareGraphicsExtensions{.pdf,.jpg,.png}
\graphicspath{{./images/}}

\newcolumntype{C}{X<{\centering}}

\definecolor{lightgreen}{rgb}{0.67, 0.88, 0.69}
\definecolor{darkgreen}{rgb}{0,0.55,0}
\definecolor{linkcolor}{rgb}{0,0,.65}

\newcommand\Caption[3][]{\caption[#2]{\label{#1}\textsc{#2}. \it\small#3}}

\renewcommand\sec[1]{Sec.~\ref{sec:#1}}
\newcommand\fig[1]{Fig.~\ref{fig:#1}}

\newcommand\tab[1]{Tab.~\ref{tab:#1}}

\usepackage[pagenumbers]{cvpr}
\usepackage{fancyhdr}

\usepackage[pagebackref=true,breaklinks=true,colorlinks,bookmarks=false]{hyperref}

\begin{document}

\title{Quo Vadis RankList-based System in Face Recognition?}

\author{Xinyi Zhang \qquad Manuel G\"unther\\[.5ex]
Department of Informatics, University of Zurich\\
Andreasstrasse 15, CH-8050 Zurich\\
{\tt\small \{xinyi.zhang,manuel.guenther\}@uzh.ch}
}

\maketitle
\fancyhead[C]{\small This is a pre-print of the original paper accepted for presentation at the International Joint Conference on Biometrics (IJCB) 2024.}
\thispagestyle{fancy}

\begin{abstract}

Face recognition in the wild has gained a lot of focus in the last few years, and many face recognition models are designed to verify faces in medium-quality images.
Especially due to the availability of large training datasets with similar conditions, deep face recognition models perform exceptionally well in such tasks.
However, in other tasks where substantially less training data is available, such methods struggle, especially when required to compare high-quality enrollment images with low-quality probes.
On the other hand, traditional RankList-based methods have been developed that compare faces indirectly by comparing to cohort faces with similar conditions.
In this paper, we revisit these RankList methods and extend them to use the logits of the state-of-the-art DaliFace network, instead of an external cohort.
We show that through a reasonable Logit-Cohort Selection (LoCoS) the performance of RankList-based functions can be improved drastically.
Experiments on two challenging face recognition datasets not only demonstrate the enhanced performance of our proposed method but also set the stage for future advancements in handling diverse image qualities.

\end{abstract}

\section{Introduction}
Face recognition has been one of the first successful approaches of machine learning.
Already traditional face recognition methods surpass human performance, as long as the capturing conditions can be controlled \cite{otoole2007surpass}.
However, they still fail in more complex scenarios \cite{guenther2016survey}, for example, when trying to compare faces across viewing angles or under different illumination.
The main reason for the former limitations is that the features extracted by traditional approaches are often localized, and the appearance of the features differed strongly from frontal to profile faces.

Based on the observation that it might not be a good idea to directly compare incomparable features, M\"uller \etal\cite{mueller2007unenrolled,mueller2013invariant} and Schroff \etal\cite{schroff2011pose} proposed to indirectly compare faces via RankLists, aka., the Doppelganger approach.
For example,  M\"uller~\etal~\cite{mueller2007unenrolled} applied this technique for frontal-to-profile face recognition, where a gallery is enrolled using frontal faces, while probe images are taken in (half-)profile view.
This process relies on defining two splits of a cohort dataset, one including frontal and one profile faces, both capturing the same identities.
A visual depiction of this process is given in \fig{rl1}.
Now, frontal faces are compared to everyone in the frontal cohort, building a list of similarity scores, while profile faces are compared to the profile cohort.
These two similarity score lists can then be compared using dedicated similarity functions, for example, by turning them into rank lists first \cite{mueller2013invariant}.
Interestingly, M\"uller~\etal~\cite{mueller2007unenrolled} used two related yet different algorithms to compare frontal-to-frontal-cohort and profile-to-profile-cohort faces.
They showed that their indirect approach outperformed the other traditional approaches that rely on direct comparisons across pose and across illumination by a large margin \cite{mueller2013invariant}.

The introduction of deep learning into face recognition \cite{schroff2015facenet,parkhi2015deep} enabled the usage of large amounts of available facial image data for training a deep model \cite{wang2021survey}.
The collection of large image datasets including diverse viewing conditions of faces \cite{cao2018vggface2,zhu2021webface260m}, together with the development of specialized loss functions that produce discriminative features \cite{wang2018cosface,deng2019arcface,kim2022adaface}, enabled the deep learning models to compare faces across pose \cite{pereira20228years}.
Also, it was indicated \cite{mylaeus2022bachelor} that the traditional RankList-based approaches using deep features similarities for comparing faces no longer have a benefit over the direct comparison of frontal and profile faces.

The success of deep-learning approaches led to the collection of more complex face recognition benchmarks, such as the IARPA Janus Benchmark (IJB) series \cite{maze2018ijbc}, the Cross-Pose LFW dataset \cite{zheng2018cross}, and lately the Biometric Recognition and Identification at Altitude and Range (BRIAR) program \cite{cornett2023briar}.
While the IJB-C dataset has mainly focused on the same type of images -- different illuminations, facial expressions, and yaw angles -- as present in the large-scale training datasets, BRIAR goes beyond that and requires comparison of faces with different pitch angles, or with different distances to the camera.
Particularly, enrollment is performed from high-quality, high-resolutions and well-illuminated faces, whereas probe images are taken at a distance or from UAVs, and might contain severe atmospheric disturbances and pitch angles.
Since such conditions are not included in the large-scale training datasets -- and training with artificial atmospheric perturbations is not sufficient to improve recognition drastically \cite{robbins2024daliid} -- purely training-based approaches are (currently) not sufficient to tackle these problems.

Now, we are back to the initial conditions: we have a benchmark to compare faces that deep learning methods are not able to compare directly.
Therefore, in this paper we try to exploit RankList-based methods using deep features in order to improve face recognition across pitch and distance.
We highlight two issues hindering the applicability of RankList-based approaches.
To mitigate these issues, we extend the traditional RankList approaches to make use of the logits of face recognition networks, and propose the Logit-Cohort Selection (LoCoS) algorithm for indirect face matching.
We show that indirect matching relies mostly on the most similar \emph{and the most dissimilar} faces in the cohort.
We make use of the DALIface model \cite{robbins2024daliid} for extracting features and logits from images.
We show that LoCoS can dramatically improve over the other RankList-based methods, while at the same avoiding the definition and the explicit comparison to a cohort.
We highlight that, in principle, this method can work, and we keep the same perfect recognition performance when comparing good-quality to good-quality faces.
Due to the lack of face recognition algorithms that are specialized for comparing high-pitch to high-pitch faces, or high-atmospherics to high-atmospherics faces -- DALIface is a good start but not directly designed for such tasks -- the LoCoS approach is still inferior to the direct comparison of faces.
However, we are on a good way of closing the gap and future work will focus on designing face recognition methods for directly comparing difficult faces with each other.

\section{Related Work}
\begin{figure*}
\centering
\begin{tikzpicture}[auto, node distance=2cm, >=Stealth]

\node[rectangle] (dataset)[draw,thick,minimum width=.5cm,minimum height=3.5cm] {\rotatebox{90}{Dataset}};

\node (cohort) [rectangle, draw, right=1cm of dataset] {cohort dataset};
\node (gallery) [rectangle, draw, above=of cohort] {gallery $x_g$};
\node (probes) [rectangle, draw, below=of cohort] {probe $x_p$};

\node (cohort_gallery) [rectangle, draw,right=1cm of cohort,yshift=1cm] {cohort $X_{cg}$};
\node (cohort_probe) [rectangle, draw,right=1cm of cohort,yshift=-1cm] {cohort $X_{cp}$};

\node[rectangle] (extractor)[draw,thick,minimum width=.5cm,minimum height=3.5cm,right=4cm of cohort] {\rotatebox{90}{feature extractor}};

\node (f_g) [rectangle, draw,right=1cm of extractor, yshift=2cm] {$\phi_g$};
\node (f_cohort_g) [rectangle, draw,right=1cm of extractor,yshift=1cm] {$\Phi_{cg}$};
\node (f_cohort_p) [rectangle, draw,right=1cm of extractor, yshift=-1cm] {$\Phi_{cp}$};
\node (f_p) [rectangle, draw,right=1cm of extractor, yshift=-2cm] {$\phi_p$};

\node (fgg) at ($(f_g)!0.5!(f_cohort_g)$) {};
\node (fpp) at ($(f_p)!0.5!(f_cohort_p)$) {};
\node (sim1) [rectangle, draw,right=1cm of fgg]  {\rotatebox{90}{similarities $L_g$}};
\node (sim2) [rectangle, draw,right=1cm of fpp] {\rotatebox{90}{similarities $L_p$}};

\node (rk1) [rectangle, draw,right=1cm of sim1] {\rotatebox{90}{ranklist $\gamma_g$}};
\node (rk2) [rectangle, draw,right=1cm of sim2] {\rotatebox{90}{ranklist $\gamma_p$}};

\node (rkc) at ($(rk1)!0.5!(rk2)$) {};
\node (score) [rectangle, draw,right=1cm of rkc] {score $s$};

\draw[->] (dataset) -- (gallery);
\draw[->] (dataset) -- (cohort);
\draw[->] (dataset) -- (probes);

\draw[->,double] (cohort) -- (cohort_gallery);
\draw[->,double] (cohort) -- (cohort_probe);

\draw[->] (gallery) -- ($(extractor.west) + (0,1.4)$);
\draw[->,double] (cohort_gallery) -- ($(extractor.west) + (0,0.4)$);
\draw[->,double] (cohort_probe) -- ($(extractor.west) - (0,0.4)$);
\draw[->] (probes) -- ($(extractor.west) - (0,1.4)$);

\draw[->] ($(extractor.east) + (0,1.4)$) -- (f_g);
\draw[->,double] ($(extractor.east) + (0,0.4)$) -- (f_cohort_g);
\draw[->,double] ($(extractor.east) - (0,0.4)$) -- (f_cohort_p);
\draw[->] ($(extractor.east) - (0,1.4)$) -- (f_p);

\draw[->] (f_g) -- (sim1);
\draw[->,double] (f_cohort_g) -- (sim1);
\draw[->,double] (f_cohort_p) -- (sim2);
\draw[->] (f_p) -- (sim2);

\draw[->] (sim1) -- (rk1);
\draw[->] (sim2) -- (rk2);

\draw[->] (rk1) -- (score);
\draw[->] (rk2) -- (score);

\end{tikzpicture}

\Caption[fig:rl1]{Tradition RL process flow}{
    The figure illustrates the workflow of a face recognition system incorporating traditional ranklists. 
    The dataset provides a gallery image $x_g$, a probe image $x_p$, and a cohort set, which is further divided into several images in the cohort gallery $X_{cg}$ and cohort probe $X_{cp}$. 
    Features $\phi$ are extracted for these images firstly, where the features $\Phi$ of the cohort can be pre-computed. 
    Then these features are used to compute similarity score lists $L_g$ and $L_p$, which are converted into two ranklists ($\gamma_g$ and $\gamma_p$).
    The final similarity score $s$ is computed by a dedicated rank list similarity function.
}
\end{figure*}

\subsection{Deep Face Recognition}
In recent years, deep learning has dominated and revolutionized many fields of research, including face recognition.
In general, there exist two main directions: developing better network topologies and implementing better-suited loss functions.
Wang \etal~\cite{wang2021survey} provides a survey of algorithms, datasets, and evaluations before 2020, and more approaches have been developed since.
Most modern network architectures \cite{hu2018squeeze} include variations and improvements \cite{duta2021iresnet} of residual network architectures \cite{he2016deep}, and vision transformers \cite{sun2022part}.
The latest developed loss functions, \ie{} ArcFace \cite{deng2019arcface}, MagFace \cite{meng2021magface}, and Adaface \cite{kim2022adaface}, improve the discriminability of deep features in angular space.

Many of these networks are trained on huge amounts of data, the largest training dataset has 260 million images \cite{zhu2021webface260m}, of which typically only subsets of the data are exploited.
Since these datasets are crawled from the internet, they mainly capture celebrities, and the distribution of imaging conditions follows the standards that are used for media publishing, \ie, faces are usually of comparable high resolution and quality, and mainly show differences in illuminations, facial expressions and yaw angles -- pitch angles are typically not varied much.
To overcome the difficulty of low-quality faces, using additional augmented data has been shown to improve face recognition performance in degraded images \cite{robbins2024daliid}.
However, the lack of real data for varying pitch angles, especially capturing faces from elevated angles, currently prevents the training of deep face recognition systems that are specialized for differences in pitch angles.
To increase the number of high-pitch samples available for training, the BRIAR program \cite{cornett2023briar} is currently collecting such data.
In the meantime, we want to investigate whether traditional RankList-based methods can be used to improve performance on these tasks.

\subsection{Face Recognition with RankLists}
\label{sec:traditional}
The original idea of using RankLists or Doppelganger lists for face recognition was developed prior to the deep learning era.
Particularly, M\"uller \etal~\cite{mueller2007unenrolled,guenther2010twokinds} explored the possibility of using RankLists for comparing frontal to half-profile faces. 
To compare the frontal gallery faces with the frontal cohort faces, and profile probe faces with the profile cohort images, two separate instances of the Elastic Bunch Graph Matching algorithm \cite{wiskott1997ebgm} with different numbers of nodes were used.
Later, they extended their research to compare faces across different illuminations and resolutions \cite{mueller2013invariant}, and they proposed several functions for comparing rank lists, some of which are used in our experiments.
In a similar line of research, Schroff \etal~\cite{schroff2011pose} evaluated RankList-based methods that they termed Doppelganger lists to compare faces across pose, facial expression, and illumination.
For comparing two faces, they adopted a simple Structural Similarity Index (SSIM) \cite{wang2004image}, and developed another similarity function to compare the generated rank lists.
All RankLists comparison functions of M\"uller and Schroff indicated that lower ranks, corresponding to more similar cohort identities, should be weighted higher.
Lately, Wartmann \cite{wartmann2021bachelor} investigated the applicability of RankLists for deep face recognition across poses.
He realized that not only the most similar cohort identities are of interest, but also the least similar ones, and he proposed to adapt similarity functions accordingly.
Similarly, Mylaeus \cite{mylaeus2022bachelor} tested various similarity functions for cohort-based face recognition, and found little difference in performance when comparing high-resolution to low-resolution faces.

\subsubsection{Traditional RankList Extraction}
\fig{rl1} shows how traditional RankList methods work. 
Initially, a dataset is split into three parts, \emph{gallery} $X_g$, \textit{probe} $X_p$, and \textit{cohort} $X_c$, including images that match the conditions of the gallery ($X_{cg}$) and the probe ($X_{cp}$). 
For the comparison of a gallery image $x_g$ and a probe image $x_p$, a feature extractor is applied to obtain feature representations for the gallery ($\phi_g$) and the probe ($\phi_p$).
Additionally, features of the cohort images ($\Phi_g$ and $\Phi_p$) are obtained.
Subsequently, similarity scores are calculated in two stages. 
First, a list of similarity scores $L_g$ is computed between the gallery features $\phi_g$ and all cohort images under gallery conditions ($\Phi_{cg}$). 
Second, similarity scores ($L_p$) are calculated between the probe features ($\phi_p$) and all cohort images under probe conditions ($\Phi_{cg}$). 
Therefore, after this step, the probe and gallery image is represented through a list of similarity scores referring to subjects in the cohort. 

Next, these similarity vectors $L$ are converted into RankList $\gamma$ by giving the first rank 0 to the highest entry \ie, the most similar cohort image, rank 1 to the next highest, and so on. 
Finally, to compute the final similarity score $s$ between gallery $\gamma_g$ and probe RankLists $\gamma_p$, tailored RankList comparison functions $S(\gamma_g, \gamma_p)$ are employed, as discussed below. 


\subsubsection{Traditional RankList Comparison Functions}
To compare two RankLists, M\"uller \etal~\cite{mueller2007unenrolled} describe three criteria necessary for similarity functions: (1) identical RankLists should yield the highest similarity ($s = 1$). 
(2) More equal indices in the same rank should increase similarity. 
(3) Lower ranks should be weighted more heavily than higher ranks. 
Then they propose the following function, which we term $S_1$:


\begin{equation}
  \label{eq:m10-1}
   S_1(\gamma_g, \gamma_p)=\frac1{\hat S_1}{\sum\limits_{m=0}^{N_c-1} {(\gamma_g[m]+\gamma_p[m])}^ {-\frac14}} 
\end{equation}
%
where $N_c$ is the number of identities in the cohort, \ie, the length of the RankList.
The similarity function is normalized by dividing by the maximum possible similarity score $\hat S_1$ \cite{mueller2007unenrolled}, so that a score between 0 and 1 is produced -- since this factor is constant, it can also be left out when the final score does not need to be normalized.

\begin{figure*}
\centering
\begin{tikzpicture}
    \node at (-2,3){Gallery $x_g$};
    \node at (-2,-3){Probe $x_p$};
    \node (gallery) at (-2,-1.5) {\includegraphics[width=2cm]{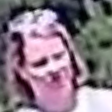}};  
    \node (probe) at (-2,1.5) {\includegraphics[width=2cm]{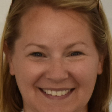}};  
    \draw[->, thick] (-1,-1.5) -- (0.5,-1.5);
    \draw[->, thick] (-1,1.5) -- (0.5,1.5);

    \node at (1, 0) (model) [rectangle, draw, minimum width=1cm, minimum height=5cm] {\rotatebox{90}{Backbone}};

    \node at (3, 1.5) (f_g) [rectangle, draw] {\rotatebox{90}{features $\varphi_g$}};
    \node at (3, -1.5) (f_p) [rectangle, draw] {\rotatebox{90}{features $\varphi_p$}};
    \draw[->, thick] (1.5,1.5) -- (f_g);
    \draw[->, thick] (1.5,-1.5) -- (f_p);

    \node at (5, 1.5) (z_g) [rectangle, draw] {$Z_g$};
    \node at (5, -1.5) (z_p) [rectangle, draw] {$Z_p$};
    \draw[->, thick] (f_g) -- (z_g) node[midway, above] {$\mathbf{W}$};
    \draw[->, thick] (f_p) -- (z_p) node[midway, above] {$\mathbf{W}$};

    \node at (6.5, 0) (idxs) [rectangle, draw] {indexes $\mathcal I$};
    \draw[->, thick] (z_g) -- (idxs); 
    

    
    \node (sim) [rectangle, draw,right=of idxs] {$s = S_{\text{LoCoS}}(Z_g, Z_p, \mathcal I)$};

    \draw[->, thick] (z_g) -- (sim); 
    \draw[->, thick] (idxs) -- (sim); 
    \draw[->, thick] (z_p) -- (sim); 



    
\end{tikzpicture}
\Caption[fig:cohort-selection]{LoCoS Cohort Selection Process Flow}{This figure displays the flow of the proposed Logit-Cohort Selection (LoCoS) method. Logits $Z_g$ and $Z_p$ are extracted from gallery $x_g$ and probe $x_p$. Indexes are selected from the gallery logits, and applied to the probe logits, via our $S_{\mathrm{LoCoS}}$ similarity function.}
\end{figure*}

Schroff \etal~\cite{schroff2011pose} discovered that the similarity value drops significantly at a certain point $k$ in the RankList, with differences becoming minimal thereafter. 
To address this, they ignored the least relevant similarities \cite{mylaeus2022bachelor}, and proposed $S_2$ to compute the similarity score between two RankLists:\footnote{The factor $k$ is not incremented by one as in the original function since the lowest rank in the lists $\gamma_g$ and $\gamma_p$ is zero in our definition.}

\begin{equation}
  \label{eq:schroff}
S_2(\gamma_g, \gamma_p)=\sum_{m=0}^{N_c-1}(k+1-\gamma_g[m])_{+} \cdot(k+1-\gamma_p[m])_{+}
\end{equation}
where $(\cdot)_+ = \max(\cdot,0)$ removes any negative value, \ie, where either gallery rank $\gamma_g[m]$ or probe ranks $\gamma_p[m]$ is higher than $k$ representing lower similarity. 
Note that due to $(\cdot)_+$, the results for values where $k > N_c$ are equivalent to those with $k = N_c$. 
Also, this function is not normalized, so values are typically much larger than 1.

Two years later, M\"uller \etal~\cite{mueller2013invariant} published a new similarity function $S_3$, which can guarantee a faster decay and therefore yields better results:
\begin{equation}
\label{eq:m13-1}
S_3(\gamma_g, \gamma_p)=\frac1{\hat S_3}{\sum\limits_{m=0}^{N_c-1} \lambda^{\gamma_g[m]+\gamma_p[m]}}
\end{equation}
using the constant normalization factor $\hat S_3$ \cite{mueller2013invariant}.
They indicated that parameters $0<\lambda<1$ should be chosen rather on the larger end, and the proposed $\lambda=0.99$.

Finally, Mylaeus \cite{mylaeus2022bachelor} investigated the use of standard comparison functions including Euclidean distance or cosine similarity, which he applied directly to the similarity vectors $L$, but his results indicate that ranklist-based methods perform on par or better.

\section{Approach}

The aim of this paper is to assess the utility of RankList-based methods in low-resolution, low quality and -- through evaluating on the BRIAR data -- high-pitch recognition conditions.
We refer to high-quality gallery-enrollment images with HQ, and low-quality low-resolution probe images with LR.
To improve the cohort size and content as compared to traditional methods discussed in \sec{traditional}, we provide a new algorithm -- \textbf{Logit-Cohort Selection} (\textit{\textbf{LoCoS}}), which can use the predictions of the face recognition network's last layer to compute similarity.


\subsection{Logit-Cohort Selection}
The above approaches using cohort datasets to compute RankLists have two major issues.
First, a large set of cohort images is required where each identity is captured in all gallery and probe conditions, which might be difficult to obtain.
Second, to compute the similarity list $L$ and the final RankList $\gamma$, it is necessary to calculate feature similarity scores $N_c$ plus one RankList comparison score, which could be excessively expensive.

To avoid the usage of a cohort dataset altogether and increase the size of the cohort at the same time, we propose our Logit-Cohort Selection (LoCoS) model, which we depict in \fig{cohort-selection}.
To understand this method, we first need to introduce the way that face recognition networks are typically trained.
An image $x$ of identity $c$ is input to the network, which extracts deep features $\varphi$.
Then, these features are projected into logit space $Z=\mathbf W \cdot \varphi$ via matrix $\mathbf W$, and the logits are used through softmax-based loss functions \cite{parkhi2015deep} to train the network.
A column $\mathbf W_c$ can be interpreted as the mean representation for training identity $c$ \cite{wen2016centerloss}, especially when normalized as required by more recent loss functions \cite{deng2022arcface,meng2021magface}:
\begin{equation}
    \label{eq:logit} z_c = \cos(\mathbf W_c, \varphi) 
                         = \frac{\mathbf W_c \cdot \varphi}{\|\mathbf W_c\| \cdot \|\varphi\|}
\end{equation}
Typically, the final matrix $\mathbf W$ is thrown away after training.
However, as we can interpret the logit $z_c$ as the similarity of the sample $x$ to training subject $c$, we can make use of the logits to define our cohort, so that we can interpret $Z$ as our list of similarities $L$.

On top of this logit list, any traditional RankList comparison method, as discussed in \sec{traditional}, can be applied theoretically, by simply turning the logit list $L$ into a RankList $\gamma$.
However, since logit lists are huge, the according template sizes to store for the gallery would be tremendous, and the computation of similarities takes a long time.
Therefore, a better strategy for selecting and storing a subset of the logits needs to be applied, and a better strategy for comparing similarity lists needs to be engineered.

Here, we define three strategies, each of which is reducing the number of employed logits.
As a baseline, we take the first logits, which correspond to the first $K=500$ subjects in the training dataset -- this can be regarded as a random selection approach.
Following the intuition of M\"uller \etal \cite{mueller2007unenrolled,mueller2013invariant} and Schroff \etal~\cite{schroff2011pose}, we select the $K$ highest logit values corresponding to the $K$ most similar identities in the training set. 
Here we select these $K$ indexes $\mathcal I$ based on the sorted similarities/logits of the gallery $Z_g$, and we apply the exact same $\mathcal I$ to the probe $Z_p$.

After concatenating these, any traditional RankList comparison function can be applied.
However, since logit scores are the products of a deep network, and since we have learned that deep features are best compared via cosine similarity \cite{liu2017sphereface}, we employ the faster and easier-to-implement cosine similarity (using python-indexing notation):
\begin{gather}
    S_{LoCoS}(Z_g, Z_p, \mathcal I) = \cos(Z_g[\mathcal I_{[:T]}], Z_g[\mathcal I_{[:T]}]) \nonumber\\[.5ex]  + \cos(Z_g[\mathcal I_{[-B:]}], Z_p[\mathcal I_{[-B:]}])
    \label{eq:LoCoS}
\end{gather}
where $\mathcal I_{[:T]}$ indicates the largest $T$ indexes, and $\mathcal I_{[-B:]}$ the smallest $B$ indexes.
By default, we set $T=K$ and $B=0$ -- removing the second part of \eqref{eq:LoCoS}.
Additionally, following the indication of Wartmann \cite{wartmann2021bachelor} that the most \emph{and least} similar identities are important in the comparison, we investigate the $T=\frac K2$ largest and the $B=\frac K2$ smallest logit scores.




\section{Experiments}
\begin{table}
    \Caption[tab:methods]{Tested Methods}{This table lists the configurations for our experiments, including different types of cohorts, similarity functions and logit selection strategies.}
    \begin{tabular}{|l||c|c|c|}
        \hline
        \bf Name & \bf Cohort & \bf Sim & \bf Selection\\\hline\hline
        Baseline & --- & $\cos$ & embeddings\\\hline
        Cohort + $S_i$ & external & $S_i$ & all \\\hline
        LoCoS-Random & \multirow{6}{*}{logits} & \multirow{4}{*}{$\cos$} & random 500\\
        LoCoS-T &  &  & top 500\\
        LoCoS-TB &  &  & top + bottom 250\\
        LoCoS-P &  &  & top 500 of probe\\\cline{3-4}
        LoCoS-T + $S_i$ & & \multirow{2}{*}{$S_i$} &  top 500\\
        LoCoS-TB + $S_i$ & &  &  top + bottom 250\\\hline
         
    \end{tabular}
\end{table}
\subsection{Experimental Setup}
The pre-trained network used in this paper is DaliFace \cite{robbins2024daliid}, a ResNet100 model trained with AdaFace on the WebFace4M dataset. 
Robins \etal~\cite{robbins2024daliid} implemented an atmospheric turbulence simulator for data augmentation and introduced a novel distortion loss to enhance robustness, which has been empirically demonstrated to be highly effective in low-quality scenarios. 
Particularly, DaliFace is a combined model of two networks, one trained with (LQ) and one trained without (HQ) atmospheric turbulence, and their deep feature representations are not directly comparable -- while their logit vectors are.
Additionally, we also obtained the pre-trained weights of the last layer $\mathbf W$ for both networks, which represents the logit scores for 617'970 identities of WebFace4M \cite{zhu2021webface260m}.

Our experimental setup is listed in \tab{methods}.
In most of our experiments, we rely on the atmospheric-adapted LQ network only.
The \textit{\textbf{Cohort + $S_i$}} functions implement the traditional RankList strategies using the three ranklist similarity functions introduced in \sec{traditional}.
In our \textit{\textbf{Baseline}}, we extract features $\varphi_g$ and $\varphi_p$ with the LQ network, and we compare them via cosine similarity.
In \textit{\textbf{LoCoS}}, we make use of the $T=K=500$ best features that we select on the gallery logits $Z_g$.
We also evaluate \textit{\textbf{LoCoS-P}} where we select the indexes based on the probe logits $Z_p$ -- note however that this method is of rather theoretical nature since the entire logit vector needs to be stored during enrollment since indexes are only selected during probing.
We test whether $T=B=250$ provides better recognition scores in \textit{\textbf{LoCoS-TB}}, or whether random selection is sufficient in \textit{\textbf{LoCoS-Random}}.
Finally, we replace the cosine similarity in \eqref{eq:LoCoS} by transforming the selected logit lists into rank lists and apply the traditional rank-list similarity functions, marked with \textit{\textbf{LoCoS-T(B) + $S_i$}}.

There are two datasets applied in this paper: SCface and BRIAR.
In SCface, we compare our LoCoS method with traditional methods based on cohort-datasets, while on BRIAR we compare mainly against the baseline.
In both datasets, we follow the verification evaluation for BRIAR \cite{cornett2023briar} and report the Truth Match Rate (TMR) at a specific False Match Rate (FMR) of 10$^{-3}$, and we plot the Receiver Operating Characteristics (ROC) curves.


\subsection{SCface}
\begin{figure}[t!]
\centering
\begin{tikzpicture}

\node (label1) at (0, 3) {Probe (LR)};
\node (label2) at (3, 3) {Cohort};
\node (label3) at (6, 3) {Gallery (HR)};

\node[inner sep=0pt] (img1) at (0, 1.5) {\includegraphics[width=1.5cm]{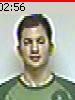}};
\node[inner sep=0pt] (img2) at (2, 1.5) {\includegraphics[width=1.5cm]{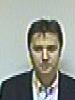}};
\node[inner sep=0pt] (img3) at (4, 1.5) {\includegraphics[width=1.5cm]{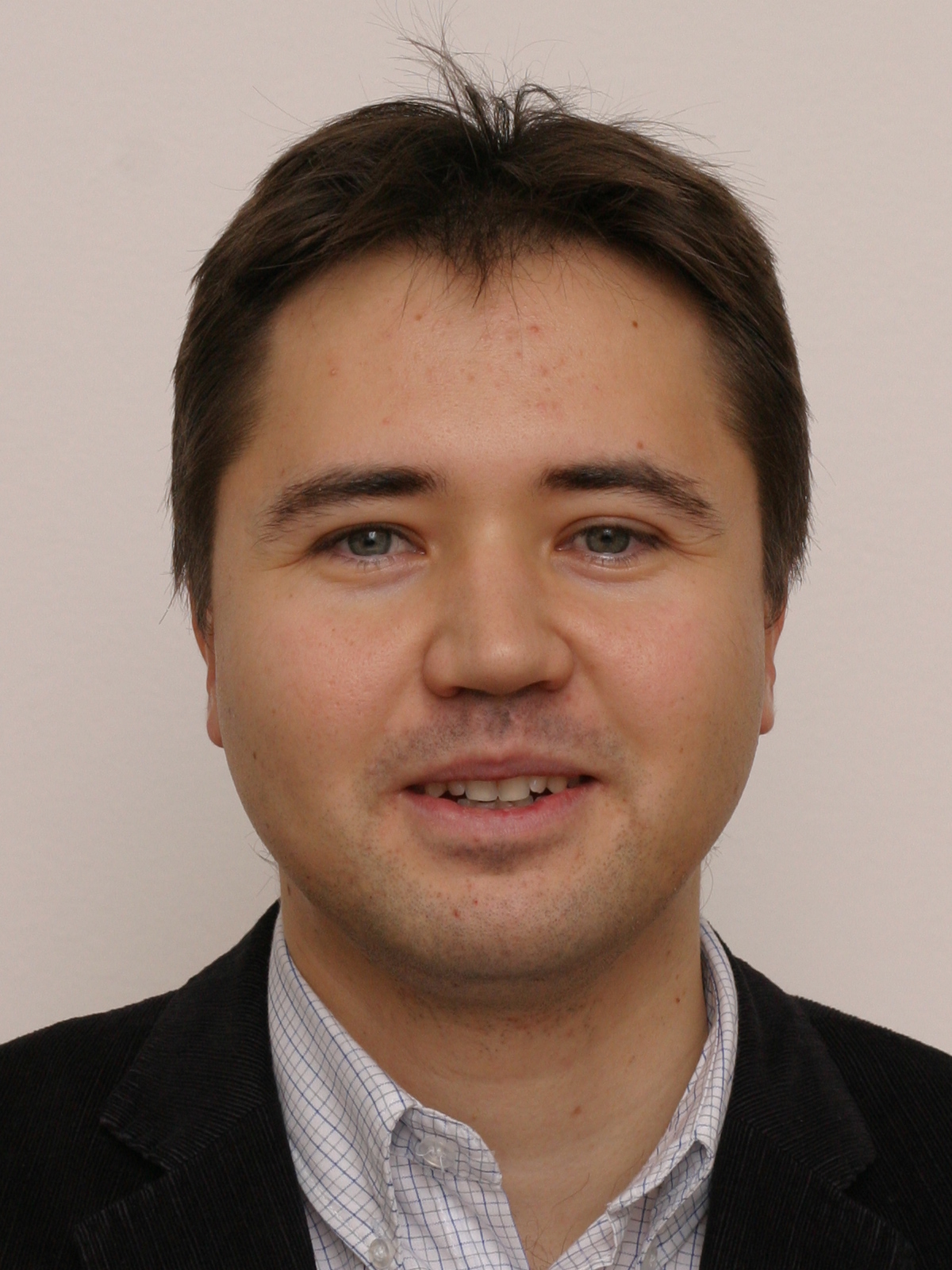}};
\node[inner sep=0pt] (img4) at (6, 1.5){\includegraphics[width=1.5cm]{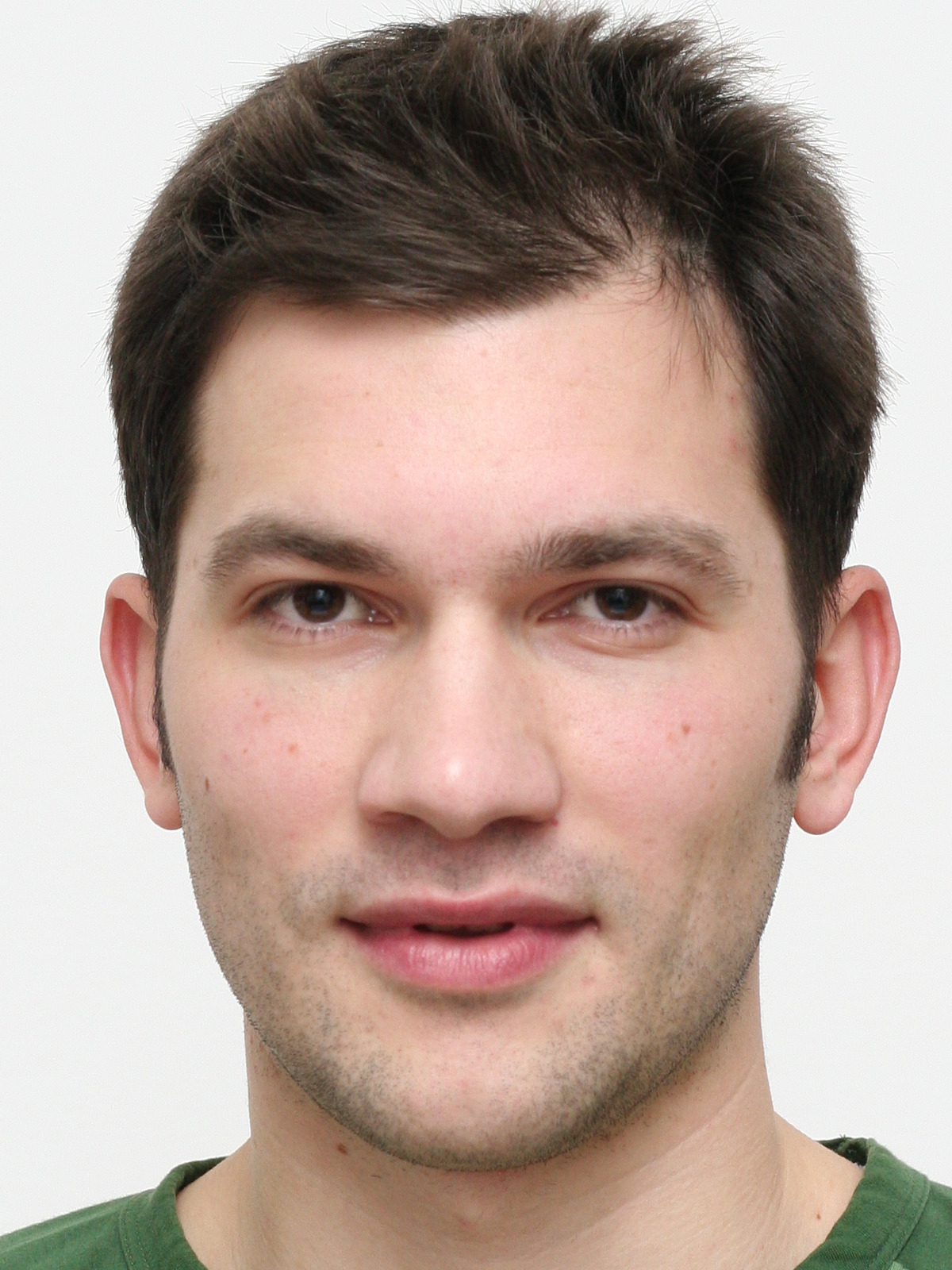}};
\node[inner sep=0pt] (img5) at (0, -1) {\includegraphics[width=1.5cm]{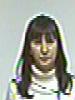}};
\node[inner sep=0pt] (img6) at (2, -1) {\includegraphics[width=1.5cm]{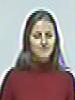}};
\node[inner sep=0pt] (img7) at (4, -1) {\includegraphics[width=1.5cm]{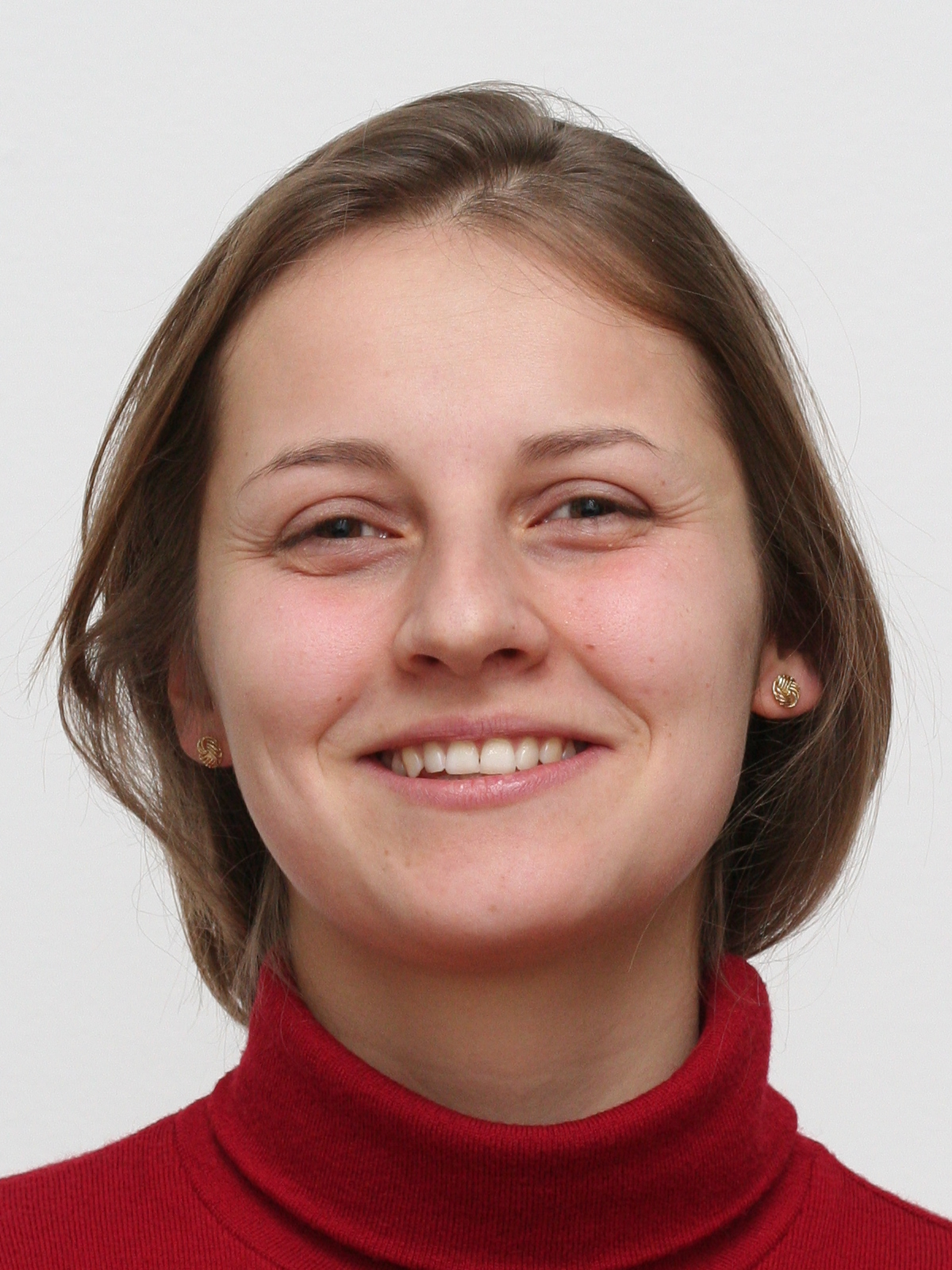}};
\node[inner sep=0pt] (img8) at (6, -1) {\includegraphics[width=1.5cm]{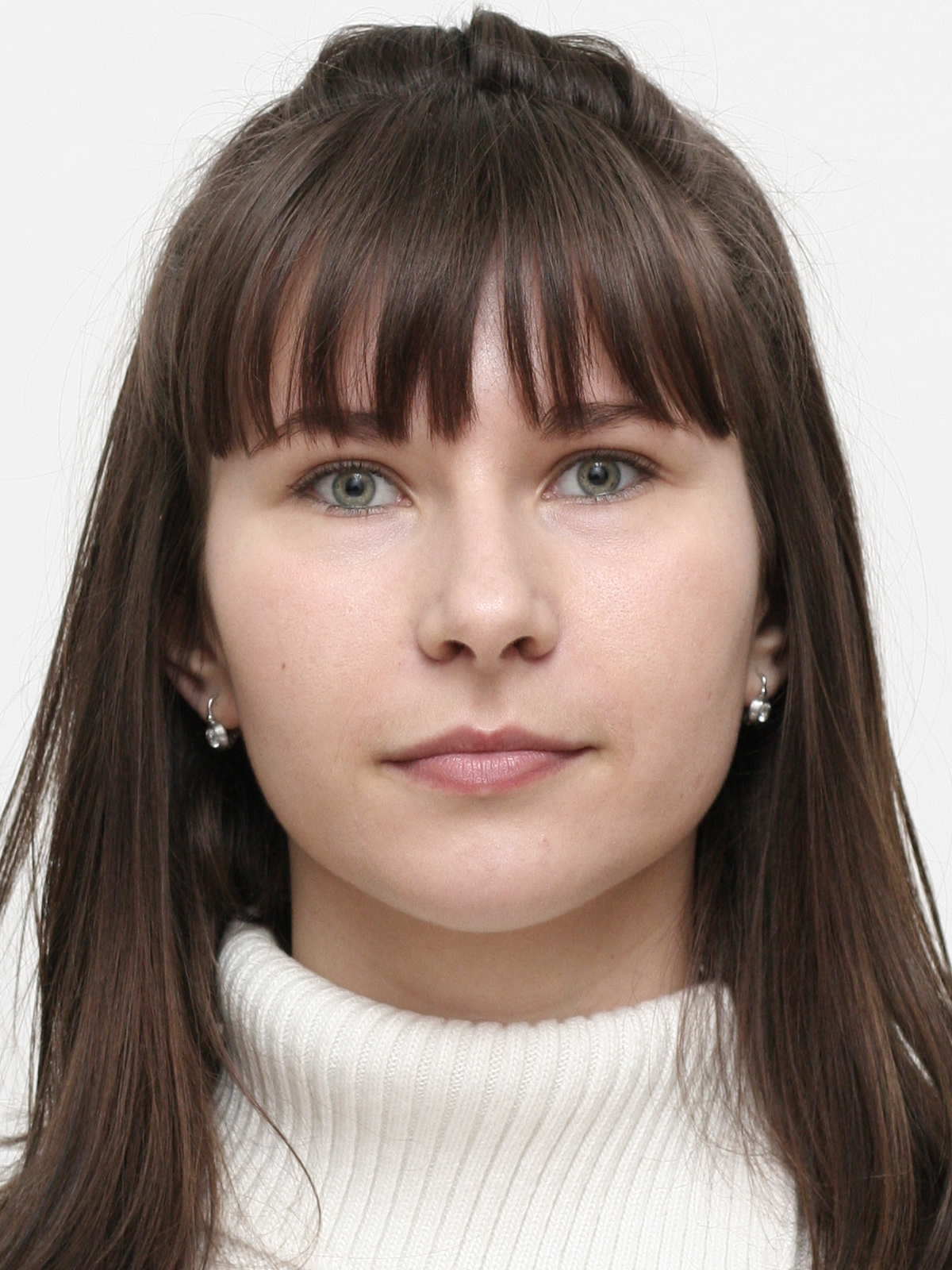}};

\draw[->] (img1.east) -- (img2.west);
\draw[->] (img1.east) -- (img6.west);

\draw[->] (img5.east) -- (img2.west);
\draw[->] (img5.east) -- (img6.west);

\draw[<-] (img3.east) -- (img4.west);
\draw[<-] (img3.east) -- (img8.west);

\draw[<-] (img7.east) -- (img4.west);
\draw[<-] (img7.east) -- (img8.west);

\draw[dotted] (img1.south) -- (img5.north);
\draw[dotted] (img2.south) -- (img6.north);
\draw[dotted] (img3.south) -- (img7.north);
\draw[dotted] (img4.south) -- (img8.north);

\end{tikzpicture}
\Caption[fig:rl-scface]{Traditional Approaches}{This figure highlights the way of defining a cohort for traditional ranklist-based methods.}
\label{}
\end{figure}
SCface \cite{grgic2011scface} comprises 4160 images sourced from 130 subjects, captured by diverse low-resolution video surveillance cameras positioned at different distances, as well as portrait-style enrollment data as seen in \fig{rl-scface}.
We apply the default evaluation protocols \cite{wallace2011intersession,guenther2016survey}, which split the probe data into three protocols: \texttt{close}, \texttt{medium}, and \texttt{far}.
Each protocol includes 44 identities for the validation set, 43 for the test set, and an additional 44 for the cohort set. 
The test set is exclusively used for face verification. 
To increase the cohort size for RankList methods, we rebuild the cohort by combining the original cohort with the validation set.

The results on the SCface dataset are provided in \tab{scface} and \fig{scface}.
When we observe the performances of three different protocols in \tab{scface}, the \textit{\textbf{Baseline}}, all \textit{\textbf{LoCoS}} variants \textit{\textbf{LoCoS-TB, LoCoS-P, LoCoS-T, LoCoS-Random}}, and combinations methods \textit{\textbf{LoCoS-TB + $S_i$}} achieve a perfect TMR of 1.0, which indicates that these methods perform optimally for protocol \texttt{close}. 
The TMRs of traditional RankList algorithms \textit{\textbf{Cohort + $S_2$}} and \textit{\textbf{Cohort + $S_3$}} based on an FMR of 0.001 are very close to 1.0, \textit{\textbf{Cohort + $S_1$}} has a worse performance. 
For the \texttt{medium} protocol, the LoCoS methods still show very competitive results: \textit{\textbf{LoCoS-P}} matches the \textit{\textbf{Baseline}}'s performance, while both the \textit{\textbf{LoCoS-TB}} and \textit{\textbf{LoCoS-T}} achieve a slightly lower TMR. 
These results are significantly better than the performance of traditional RankList methods \textit{\textbf{Cohort + $S_i$}}. 

Similarly, the \textit{\textbf{Baseline}} method maintains the highest performance in the \texttt{far} protocol. 
The \textit{\textbf{LoCoS}} methods also perform well, especially \textit{\textbf{LoCoS-T}}, followed by \textit{\textbf{LoCoS-TB}} and \textit{\textbf{LoCoS-P}} being the closest competitors to the \textit{\textbf{Baseline}}. 
These results of LoCoS methods are markedly better than those of the traditional \textit{\textbf{Cohort + $S_i$}} approaches.
It is also observed that \textit{\textbf{LoCoS + $S_i$}} variants including traditional Ranklist functions in Fig.\ref{fig:scface} outperform \textit{\textbf{Cohort + $S_i$}} across all protocols, closely approximating the \textit{\textbf{Baseline}}, particularly in the \texttt{medium} and \texttt{far} protocols. 

\begin{table}[t]
    \centering
    \Caption[tab:scface]{Results on SCface}{This table shows TMR@FMR $=10^{-3}$ for each methods of different protocols for SCface.}
    \begin{tabular}{|l||c|c|c|}
    \hline
        Method & Close & Medium & Far \\ \hline\hline
        Baseline & 1.0 & 0.995 & 0.953\\ \hline\hline
        Cohort + $S_1$   & 0.442 & 0.256 & 0.070\\ \hline
        Cohort + $S_2$   & 0.953 & 0.395 & 0.070 \\ \hline
        Cohort + $S_3$   & 0.930 & 0.442 & 0.116 \\ \hline\hline
        LoCoS-TB  & 1.0 & 0.991 & 0.837 \\ \hline
        LoCoS-P   & 1.0 & 0.995 & 0.856 \\ \hline
        LoCoS-T   & 1.0 & 0.991 & 0.907 \\ \hline
        LoCoS-Random   & 1.0 & 0.953 & 0.726 \\ \hline\hline
        LoCoS-T + $S_1$   & 0.874 & 0.586 & 0.205 \\ \hline
        LoCoS-T + $S_2$   & 0.833 & 0.6 & 0.214 \\ \hline
        LoCoS-T + $S_3$   & 0.898 & 0.623 & 0.223 \\\hline\hline
        LoCoS-TB + $S_1$   & 1.0 & 0.977 & 0.814 \\ \hline
        LoCoS-TB + $S_2$   & 1.0 & 0.995 & 0.842 \\ \hline
        LoCoS-TB + $S_3$   & 1.0 & 0.995 & 0.819 \\\hline
    \end{tabular}
\end{table}

\begin{figure}[t!]
    \centering
    \subfloat[Protocol -- \textit{\textbf{close}}\label{fig:sub1}]{
        \includegraphics[width=.9\linewidth]{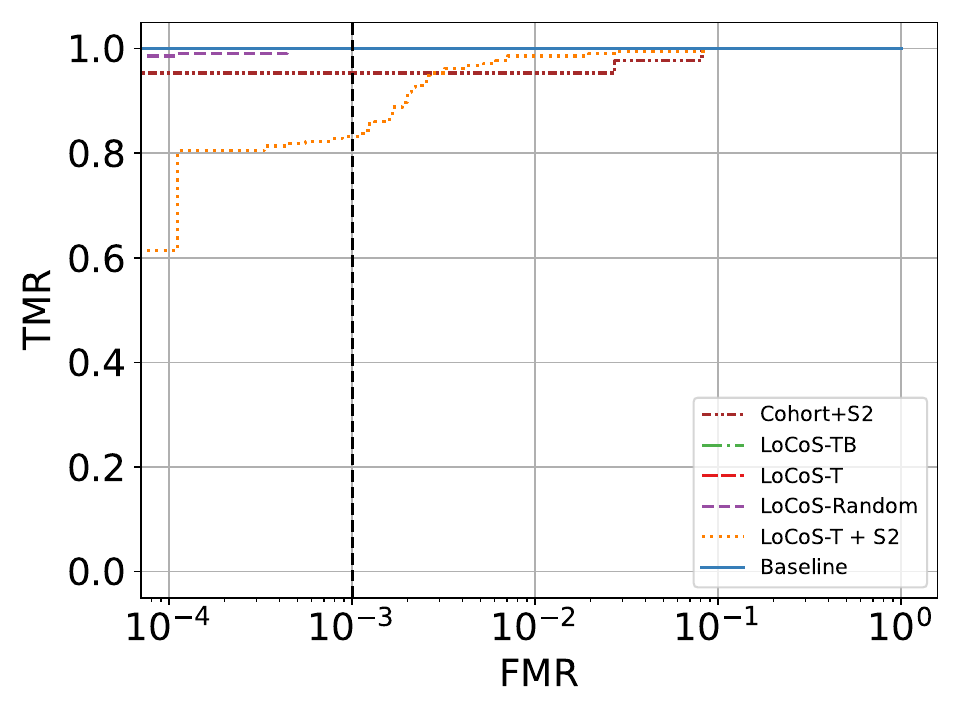}
    }
    
    \subfloat[Protocol -- \textit{\textbf{medium}}\label{fig:sub2}]{
        \includegraphics[width=.9\linewidth]{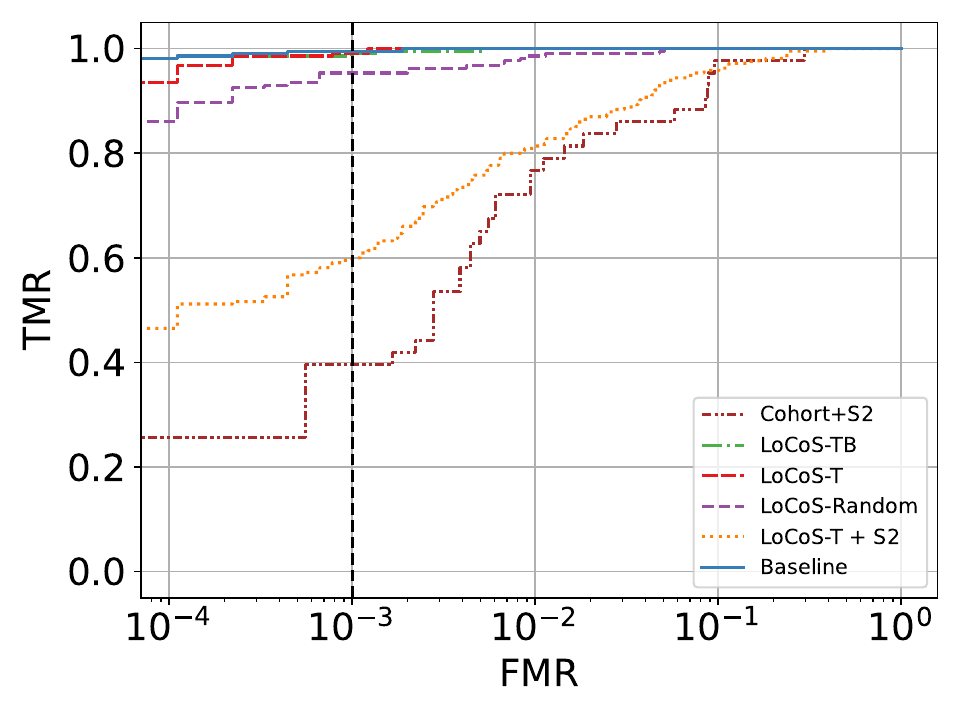}
    }
    
    \subfloat[Protocol -- \textit{\textbf{far}}\label{fig:sub3}]{
        \includegraphics[width=.9\linewidth]{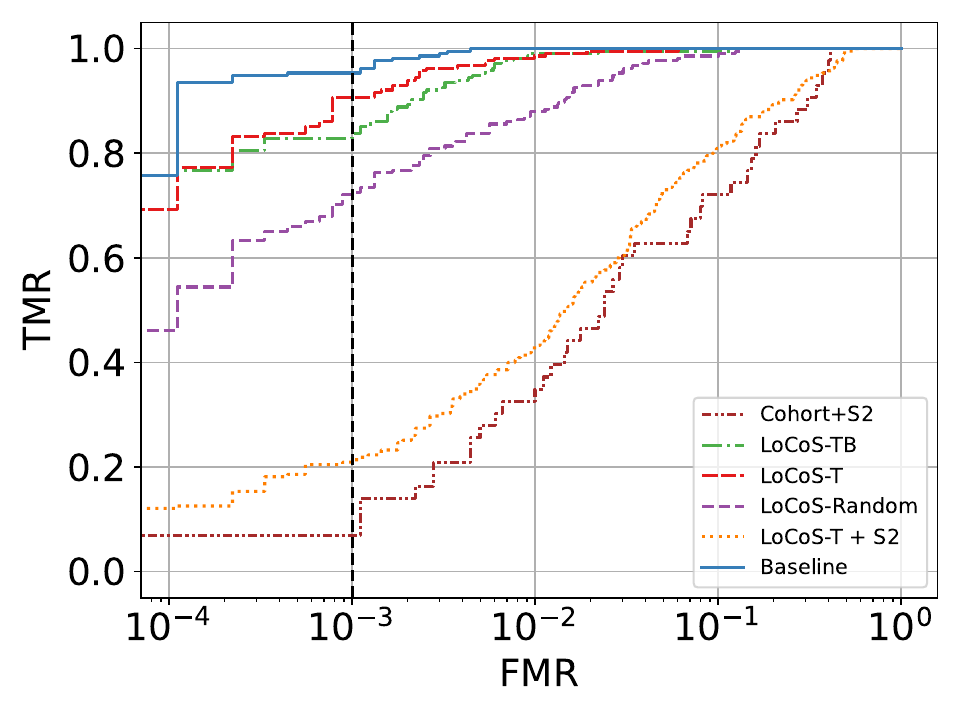}
    }
    \Caption[fig:scface]{ROC of Dataset SCface}{
        This figures shows the ROC of each protocol. At FMR=$10^{-3}$ (black dot line), higher TMR indicates better performance.
    }
\end{figure}

When we use the best \textit{\textbf{LoCoS-T + $S_i$}} with traditional RankList similarity functions, it is surprising that these performances drop extremely in protocols \texttt{medium} and \texttt{far} in \fig{scface}. 
To investigate the reason, the different combinations of LoCoS with traditional RankList techniques are applied to the protocol \texttt{far}, as we expect the biggest impact on that protocol. 
The results are presented in \fig{scface-far-diff}, revealing distinct performance differences between the combination methods of \textit{\textbf{LoCoS-T + $S_i$}} and \textit{\textbf{LoCoS-TB + $S_i$}}. 
The basic idea of traditional RankList is lower ranks should be weighted more heavily than higher ranks. 
Apparently, higher ranks coming from the non-similar subjects also impact the final FR system, while the ordering in intermediate scores is too arbitrary to improve recognition \cite{mylaeus2022bachelor}.  

We also show that the combination of two different networks for extracting gallery and probe logit lists is feasible, by using the HQ network to extract HQ gallery images, and the LQ network on LQ probe images. 
It is clear that although there is a decrease in all the methods in \fig{scface-far-sep}, it still works much better than (\textit{\textbf{Baseline (Sep)}}) which compares deep features from two networks that were trained independently; this proves the comparability of logit lists from different models. 
Since the DaliFace LQ network is still trained to compare HR with LR images, and not only LR with LR faces, an improvement could be expected when a model is trained for the latter LR to LR comparison task. 


\begin{figure}[t]
\begin{center}
   \includegraphics[width=0.8\linewidth]{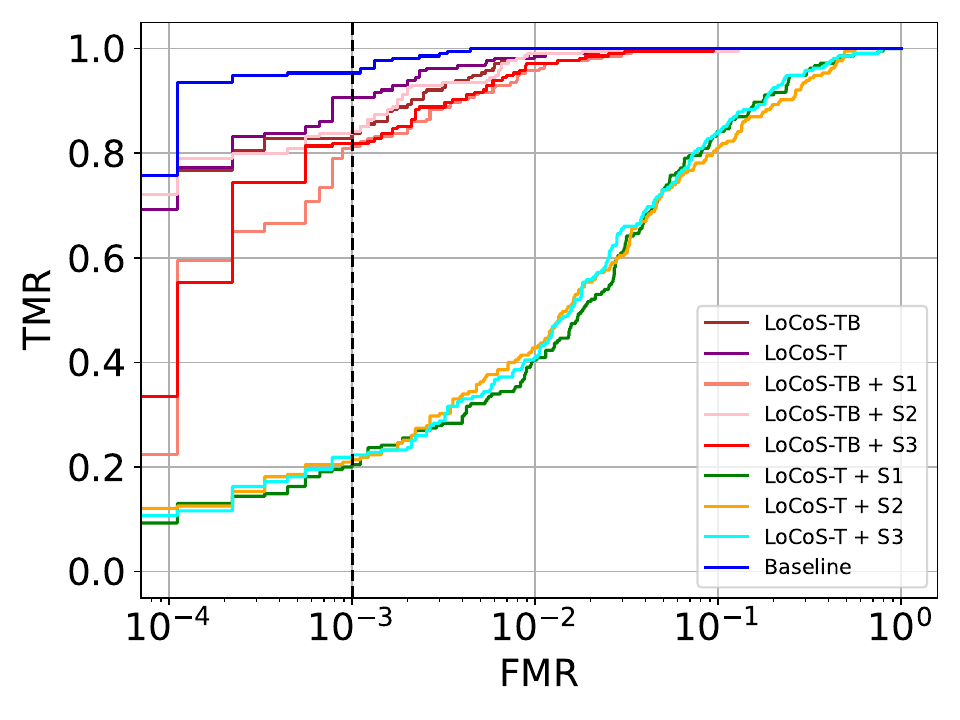}
\end{center}
\Caption[fig:scface-far-diff]{LosCoS-T vs. LosCoS-TB}{This figure shows ROC curves on the SCface dataset, protocol FAR, when using LosCoS-T or LosCoS-TB with traditional techniques.}
\end{figure}

\begin{figure}[t]
\begin{center}
   \includegraphics[width=0.8\linewidth]{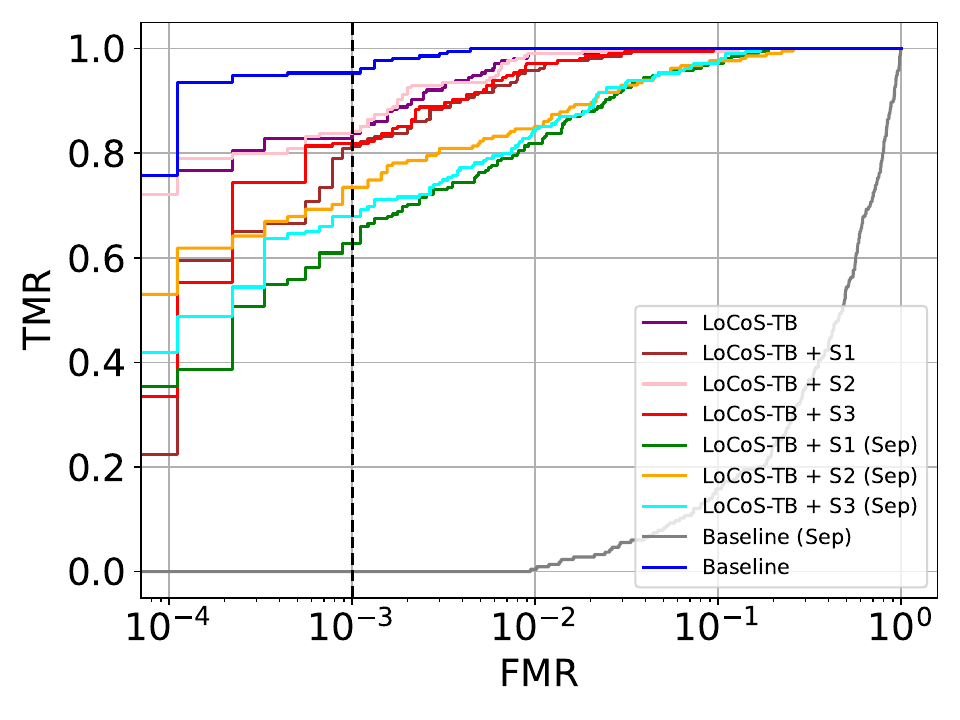}
\end{center}
\Caption[fig:scface-far-sep]{One vs. Two Networks}{This figure shows ROC curves on the SCface dataset, protocol FAR, when extracting logits with the same or with separate networks.}
\end{figure}
\subsection{BRIAR}
The second dataset we use is BRIAR \cite{cornett2023briar}.
It is designed to advance face, gait, and whole-body biometric recognition under challenging and uncontrolled conditions in full-motion video. 
The dataset addresses critical needs in security, forensics, and intelligence applications by providing data captured under extreme conditions, such as long distances, high angles, and various atmospheric perturbations.
There are in total 4 probe sets comprising two types of quality: \emph{crtl} is high(er) quality, and \emph{trt} is low-quality videos. 
It has two different sets of data: \emph{FaceIncluded} contains faces of sufficient size, while \emph{FaceRestricted} does not. 
To evaluate our RankList methods on LR face recognition, only \emph{FaceIncluded-trt} is used in this paper.
We make use of the detected faces as explained by Du \etal~\cite{du2023doers}, and we select the probe face with the largest detection score per probe video.
Our evaluation is performed on the BTS-4.2.1 evaluation protocol.
Since this protocol does not provide a specific definition of cohort, and based on the disastrous performance of the cohort dataset-based functions, we here only compare our LoCoS variations.
We divide probe \texttt{P} into four parts to evaluate our method at various distances between the camera and the subject. 
\texttt{P1} contains faces captured at close range. 
\texttt{P2} and \texttt{P3} include videos recorded from distances below and above 300 meters, respectively. 
\texttt{P4} consists of videos captured by UAVs.
Finally, protocol \texttt{P} combines all four sets, and is identical with the original protocol.

\begin{figure}[t]
\begin{center}
   \includegraphics[width=0.8\linewidth]{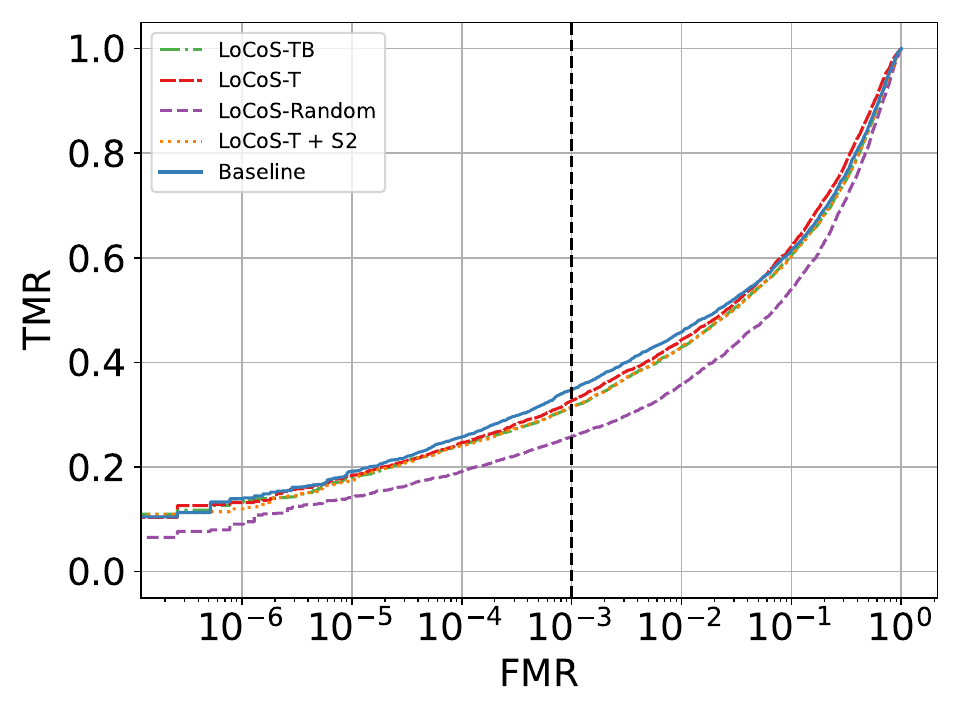}
\end{center}
\Caption[fig:briar]{ROC on BRIAR}{This figure shows the ROC of the BRIAR dataset, protocol P, indicating FMR $=10^{-3}$ with a black dotted line. }
\end{figure}


\begin{table}
    \Caption[tab:briar]{Results on BRIAR}{This table lists the verification performance (TMR@FMR=$10^{-3}$) of our proposed methods on different protocols of the BRIAR dataset.}
    
    \centering
    \small
    \begin{tabular}{|l||c|c|c|c|c|c|}
        \hline
        Method & \textit{P} & \textit{P1} & \textit{P2} & \textit{P3} & \textit{P4} \\ \hline

        Baseline & 0.347 & 0.445 & 0.368 & 0.338 & 0.135\\ \hline
        LoCoS-TB & 0.315 & 0.420 & 0.322 & 0.303 & 0.129\\ \hline
        LoCoS-P & 0.314 & 0.421 & 0.324 & 0.301 & 0.114 \\ \hline
        LoCoS-T & 0.327 & 0.427 & 0.332 & 0.318 & 0.123 \\ \hline
        LoCoS-Random & 0.258 & 0.365 & 0.242 & 0.243 & 0.079 \\ \hline
        LoCoS-TB + $S_1$ & 0.300 & 0.412 & 0.300 & 0.285 & 0.114  \\ \hline
        LoCoS-TB + $S_2$ & 0.316 & 0.415 & 0.324 & 0.304 & 0.117  \\ \hline
        LoCoS-TB + $S_3$ & 0.310 & 0.414 & 0.317 & 0.298 & 0.129  \\ \hline
    \end{tabular}
\end{table}

The results on the BRIAR dataset are given in \tab{briar} and \fig{briar}.
Although the \textit{\textbf{Baseline}} method sets the benchmark with a TMR of 0.347 for the protocol \texttt{P} for FMR $=10^{-3}$, it is followed by \textit{\textbf{LoCoS-T}} and \textit{\textbf{LoCoS-TB}}, which perform very closely, with TMRs of 0.327 and 0.315 each, demonstrating a minimal performance gap from the \textit{\textbf{Baseline}} from \tab{briar}. 
\textit{\textbf{LoCoS-P}} shows similar effectiveness, resulting in 0.314, while another variant \textit{\textbf{LoCoS-Random}} performs worse, only with the TMR of 0.258. 
\textit{\textbf{LoCoS-TB + S2}} demonstrates competitive performance among the combination methods with a TMR of 0.316. 
Finally, the TMRs of \textit{\textbf{LoCoS-TB + S1}} and \textit{\textbf{LoCoS-TB + S3}} are lower.

\fig{briar} illustrates the TMR versus FMR for various methods in the \texttt{P} protocol on the BRIAR dataset.
When the FMR $=10^{-3}$, \textit{\textbf{LoCoS-TB}} (green dash-dot line) and \textit{\textbf{LoCoS-T}} (red dashed line) closely follow the \textit{\textbf{Baseline}}, shown by the blue line with the best performance, particularly at lower FMR values, indicating their competitive performance. 
\textit{\textbf{LoCoS-T + S2}} (orange dashed line with circles) also demonstrates strong performance, nearly matching the \textit{\textbf{Baseline}} and  \textit{\textbf{LoCoS-T}}. 
In contrast, \textit{\textbf{LoCoS-Random}} (purple dashed line) performs significantly worse.

From \tab{briar}, it is apparent that as the distance between the camera and the subject increases (from protocols \texttt{P1} to \texttt{P4}), all the methods’ performance decreases dramatically, including the \textit{\textbf{Baseline}}, reducing from 0.445 to 0.135. 
Meanwhile, the performance of the \textit{\textbf{LoCoS}} variations becomes closer to that of the \textit{\textbf{Baseline}}. 
Specifically, for \texttt{P4} (faces from UAVs), the TMR values of the \textit{\textbf{LoCoS}} methods are generally more similar to the \textit{\textbf{Baseline}}, 0.006 gap compared to the gap of 0.018 for \texttt{P1} and 0.044 for \texttt{P2}, which indicates that our \textit{\textbf{LoCoS}} methods maintain competitive and stable performance at UAV conditions, while far distances are already captured well in the DaliFace LR features.


\section{Conclusion}
In this paper, we have reviewed the utility of ranklist-based methods for face recognition under conditions that are not well-preserved in current deep learning-based face recognition systems.
Particularly, we investigated very-low-resolution face recognition in SCface, as well as recognition of faces through large distances or with high-pitch angles in BRIAR.
For feature extraction, we relied on the state-of-the-art DaliFace \cite{robbins2024daliid} network, which provided excellent performance on SCface, and stable performance on BRIAR.

When applying the traditional cohort-based algorithms, large drops in performance were observed, which we partially explain by the unreasonably small cohort in SCface.
Subsequently, we extended the ranklist idea to make use of the logits of a network as representatives of cohort similarities.
With a careful selection of reliable parts of the logit vectors, we observed improved verification performance over methods without logit-cohort selection (LoCoS), using either a cosine similarity between selected logits, or dedicated rank list comparison functions.
We also indicated that it is generally possible to make use of different networks for gallery enrollment and probe extraction, when both were trained on the same identities.
However, none of our approaches was yet able to improve over the simple cosine based on the DaliFace deep features.

There are a few issues of our method to be mentioned.
First, here we only made use of one specific network, DaliFace, but initial experiments with other networks show related conclusions, but at lower absolute performance scores.
Second, the logit extraction requires the huge projection matrix $\mathbf W$ in \fig{cohort-selection} to be stored, which is typically thrown away -- therefore, we cannot use other publicly available pre-trained networks.
Third, our current cohort selection techniques are rather rough by simply selecting a fixed number of top and bottom logits.
Future work should investigate more clever and possibly sample-adapted selections, or more adapted rank list similarity functions, which might finally provide the boost to exceed baseline performance.
Fourth, in the BRIAR dataset, we only make use of a single probe face per video, while recent approaches show improved performance by integrating several frames per probe video \cite{jawade2023conan,kim2022cluster}.
Investigating the utility of such methods in our chort selection strategy would be an interesting future step.
Fifth, when incorporating two different networks for gallery and probe, these two DaliFace networks were trained in a very related way, which might have provided an optimistic view on this aspect. 
Finally, the networks were not really designed to handle high-pitch angles, or were able to extract stable features/logits for people at a distance.
Future work could make use of algorithms better suited for comparing low-quality to low-quality faces, for example, based on facial attributes \cite{rudd2016moon}.

\section*{Acknowledgements}
\vspace*{-1ex}{\fontsize{8}{8}\selectfont\noindent
This research is based upon work supported in part by the Office of the Director of National Intelligence (ODNI), Intelligence Advanced Research Projects Activity (IARPA), via 2022-21102100003.
The views and conclusions contained herein are those of the authors and should not be interpreted as necessarily representing the official policies,
either expressed or implied, of ODNI, IARPA, or the U.S.~Government.
The U.S.~Government is authorized to reproduce and distribute reprints for governmental purposes notwithstanding any copyright annotation therein.}\vspace*{-1ex}

{\small
\bibliographystyle{ieee}
\bibliography{text/References}
}

\end{document}